\documentclass[letterpaper, 10 pt, journal, twoside]{IEEEtran}   

\IEEEoverridecommandlockouts                              

\usepackage[pdftex]{graphicx}
\usepackage{amsmath} 
\usepackage{amssymb}  
\usepackage{subfigure}
\usepackage{multirow}
\usepackage{array,booktabs}
\usepackage{diagbox}
\usepackage{balance}
\usepackage{mathtools}
\usepackage{subfigure}
\usepackage{verbatim}
\usepackage{adjustbox}
\usepackage{algorithm}
\usepackage{algpseudocode}
\usepackage{romannum}
\usepackage{hyperref}
\hypersetup{
    colorlinks=true,
    linkcolor=black,
    citecolor=black,
    filecolor=black,
    urlcolor=blue,
}
\usepackage[table]{xcolor}
\makeatletter
\let\NAT@parse\undefined
\makeatother
\usepackage[numbers]{natbib}

\usepackage{rpm_math}
\usepackage{rpm_misc}
\usepackage{rpm_acronyms}
\usepackage{rpm_SIunits}

\usepackage{soul,color}

\usepackage{multirow}
\usepackage{soul,color}

\usepackage{caption}

\usepackage{lipsum}

\begin{document}

\title{\LARGE \bf
Ambiguity-Aware Multi-Object Pose Optimization\\for Visually-Assisted Robot Manipulation
}

\author{Myung-Hwan Jeon$^{1}$, Jeongyun Kim$^{2}$, Jee-Hwan Ryu$^{3}$, and Ayoung Kim$^{2*}$%
\thanks{$^\dagger$This work was supported by a grant (22TSRD-C151228-04) MOLIT.}%
\thanks{$^{1}$M. Jeon is with Dept. of the Robotics Program, KAIST, Daejeon, S. Korea
        {\tt\footnotesize myunghwan.jeon@kaist.ac.kr}}%
\thanks{$^{2}$J. Kim and A. Kim with the Dept. of Mechanical Engineering, SNU, Seoul, S. Korea
        {\tt\footnotesize [ayoungk, jeongyunkim]@snu.ac.kr}}%
\thanks{$^{3} $J. Ryu is with Dept. of Civil and Environmental Engineering, KAIST, Daejeon, S. Korea
        {\tt\footnotesize jhryu@kaist.ac.kr}}%
}


\maketitle

\begin{abstract}



6D object pose estimation aims to infer the relative pose between the object and the camera using a single image or multiple images. Most works have focused on predicting the object pose without associated uncertainty under occlusion and structural ambiguity (symmetricity). However, these works demand prior information about shape attributes, and this condition is hardly satisfied in reality; even asymmetric objects may be symmetric under the viewpoint change. In addition, acquiring and fusing diverse sensor data is challenging when extending them to robotics applications. Tackling these limitations, we present an ambiguity-aware 6D object pose estimation network, PrimA6D++, as a generic uncertainty prediction method. The major challenges in pose estimation, such as occlusion and symmetry, can be handled in a generic manner based on the measured ambiguity of the prediction. Specifically, we devise a network to reconstruct the three rotation axis primitive images of a target object and predict the underlying uncertainty along each primitive axis. Leveraging the estimated uncertainty, we then optimize multi-object poses using visual measurements and camera poses by treating it as an object SLAM problem. The proposed method shows a significant performance improvement in T-LESS and YCB-Video datasets. We further demonstrate real-time scene recognition capability for visually-assisted robot manipulation. Our code and supplementary materials are available at \href{https://github.com/rpmsnu/PrimA6D}{https://github.com/rpmsnu/PrimA6D}.


\end{abstract}


\section{Introduction}
\label{sec:intro}

6D object pose estimation is involved in recovering the orientation and translation of the object relative to the camera, playing a crucial role in various localization problems. In this line of study, deep learning-based methods \cite{xiang2018posecnn, peng2019pvnet, labbe2020cosypose, wang2019densefusion} presented remarkable performance on benchmark datasets, yet mainly focused on improving pose accuracy based on a single image. Many robotics applications often leverage multiple static cameras placed around the workspace or a single camera attached to an end-effector. In these configurations, predicting uncertainty becomes crucial to measure the prediction reliability in order to balance between inferences among multiple views or robot kinematics.


Unfortunately, in previous studies, researchers have focused on improving the prediction accuracy without associated uncertainty under occlusion, structural ambiguity (symmetricity), and object size. More critically, these studies require prior information on symmetricity for the prediction phase, which is arduous to obtain in reality; therefore, uncertainty-estimation-based methods could be more viable in field of robotics when such priors information is unavailable. Nevertheless, estimating the uncertainty for the estimated pose has been overlooked in the literature.

\begin{figure}[!t]
	\centering
	\includegraphics[width=1\columnwidth]{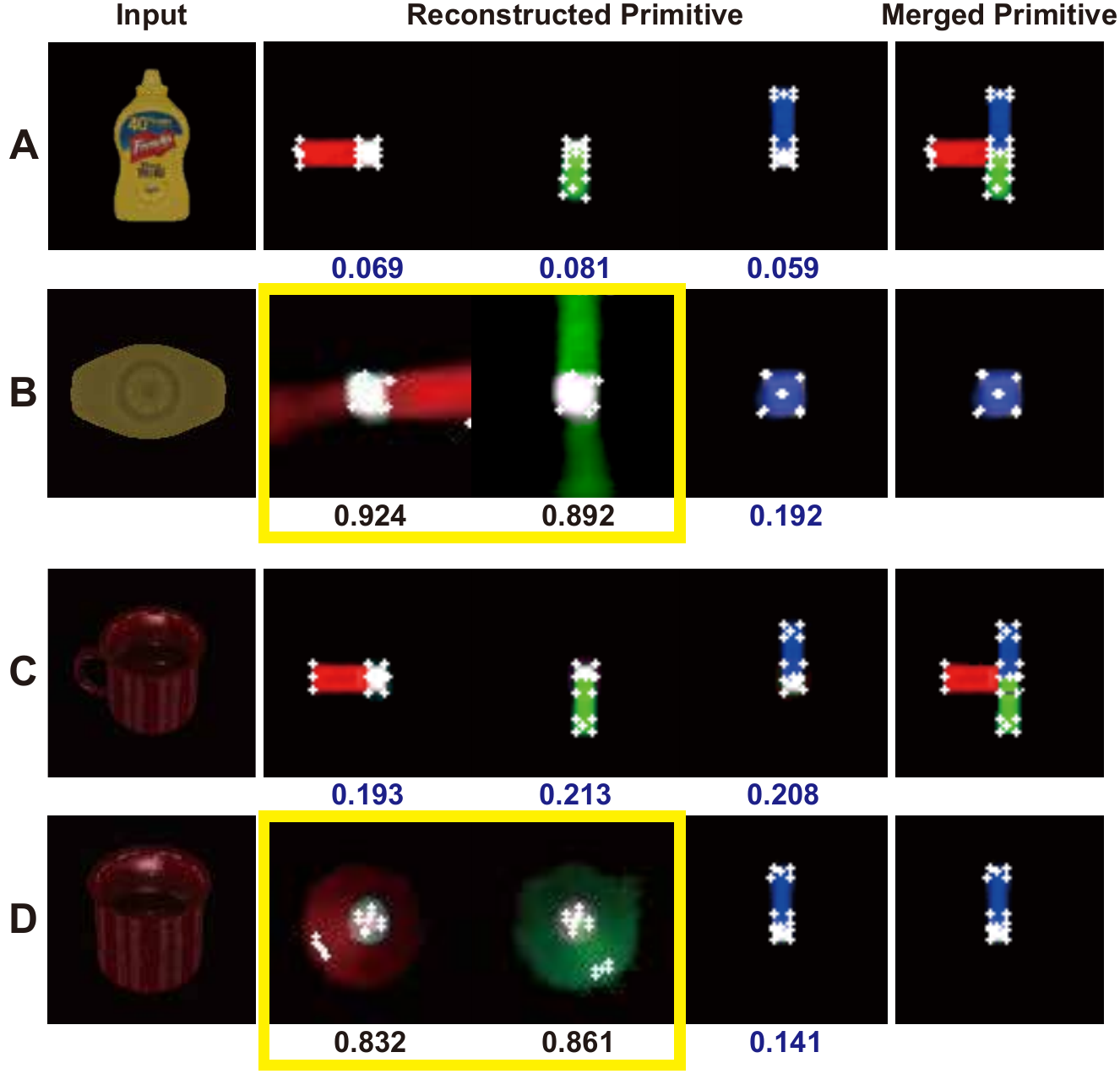}
  	\caption{Two sample cases showing varying ambiguity. An asymmetric object (A and C) may seem symmetric (B and D) depending on the viewpoint change. Unlike clearly reconstructed three rotation axes, viewpoint-induced ambiguity is captured in the reconstructed primitive axes (yellow box in B and D). We learn this induced uncertainty in an unsupervised manner to handle object ambiguity without prior information on object shape in a generic manner. }
	\label{fig:intro}
\end{figure}

Only recently, capturing uncertainty \cite{deng2021poserbpf, okorn2020learning} and including ambiguity \cite{manhardt2019explaining, jfu-2021-iros} have been reported in the literature. PoseRBPF \cite{deng2021poserbpf} employed a Rao-Blackwellized particle filter \cite{doucet2013rao} representing the distribution of the predicted pose. \citeauthor{okorn2020learning} \cite{okorn2020learning} examined uncertainties for asymmetric and symmetric objects using Bingham distribution  and histogram distribution. In \cite{manhardt2019explaining, jfu-2021-iros}, the authors indicated potential ambiguity in pose estimation and proposed leveraging multiple hypotheses but not directly estimating the uncertainty.


\begin{figure*}[!t]
	\centering
	\includegraphics[width=0.9\textwidth]{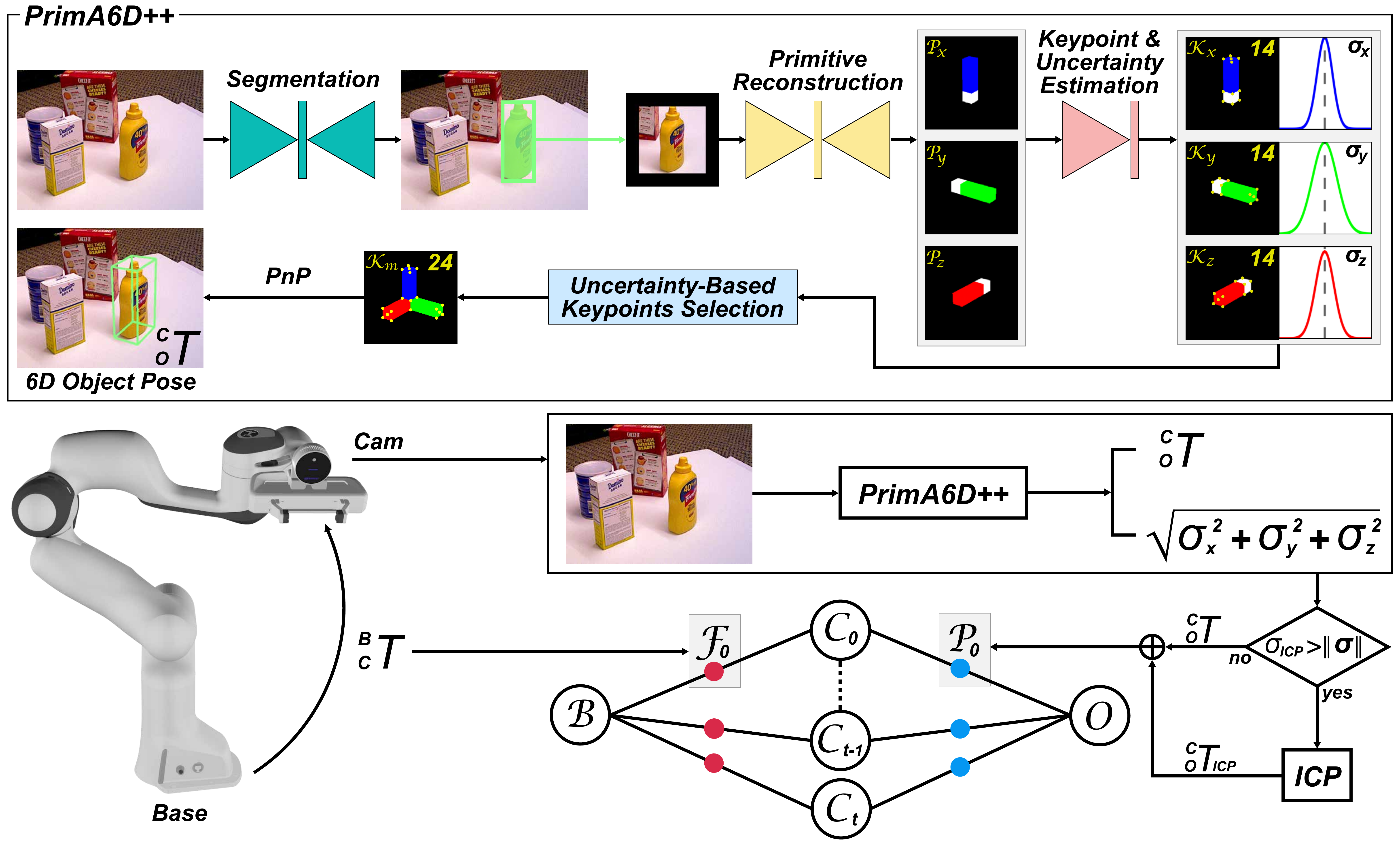}
  	\caption{The proposed network, PrimA6D++, is composed of four modules including: a segmentation network, a primitive reconstruction network, a keypoint and uncertainty extraction network, and an uncertainty-based keypoints selection module. The cropped RGB image using the object segmentation mask transforms into the three rotation axis primitive images. In each rotation axis primitive image, 14 keypoints and their associated uncertainties are estimated. To reject unreliable keypoints extracted from rotation axis primitive images, we execute the uncertainty-based keypoints selection. Lastly, the 6D object poses are refined using the factor graph optimization.
	}
	\label{fig:overview}
\end{figure*}


Unlike in previous studies, we introduce an object ambiguity-aware object pose estimation network, PrimA6D++, as a generic uncertainty estimation. Our previous work, PrimA6D \cite{jeon2020prima6d}, introduced the rotation primitive reconstruction which predicts color-coded orientation axes for robust object pose estimation, yet this method cannot handle the object ambiguity and uncertainty of the predicted pose. Extending from \cite{jeon2020prima6d}, the proposed network predicts three rotation axis primitive images, each corresponding to the orientation axis of the object (\figref{fig:intro}). In addition, the uncertainty for each rotation axis primitive image is estimated via unsupervised learning. Based on these uncertainties, we discern object ambiguity caused by shape symmetricity and occlusion by rejecting unreliable rotation axis primitive images. We further utilize the norm value of these uncertainties in the multi-object pose optimization.

In this work, a robot manipulator is equipped with an RGB-D sensor attached to the end-effector (\figref{fig:overview}). Given sequential observed sensor data from our configuration, we present a system that optimizes the multi-object poses in the workspace. Specifically, leveraging the estimated uncertainty, we formulate the problem as an object SLAM by introducing the camera pose factor and object pose factor to refine the multi-object poses with camera poses to enhance 3D scene recognition in visually-assisted robot manipulation tasks. Differing from previous methods, our method presents the following contributions.




\begin{itemize}

	\item We present a novel ambiguity-aware 6D object pose estimation network, PrimA6D++, which predicts three rotation axis primitive images and the associated uncertainties. Most existing methods require symmetricity information of an object, yet in reality, this information may not be available in advance; additionally, some asymmetric objects may be symmetric depending on the viewpoint change. The proposed uncertainty estimation is generic and does not require shape information.


	\item Leveraging the estimated uncertainty, we formulate the problem as an object SLAM by introducing the camera pose factor and object pose factor to refine the multi-object poses with camera poses. Specifically, we verify that the proposed optimization process removes the undesired jittering of the object pose estimation.

	\item We verify the meaningful performance improvement of 6D object pose estimation compared to baseline methods. The proposed method improved estimation over YCB-Video to 94.4\% in terms of ADD(-S), T-LESS to 87.8\% in terms of $e_{VSD}$, and T-LESS to 0.916 in terms of $\mathrm{AR_{mssd}}$.

\end{itemize}

\section{Related Works}
\label{sec:related}


\subsection{Single-View-Based Pose Refinement}


The 6D object pose estimation based on deep learning has achieved substantial progress, however, the accuracy is still bounded. Researchers have tried to overcome this problem with a refinement step. DPOD \cite{zakharov2019dpod} predicted the relative transformation between the input image observed via the RGB sensor and the rendered image using the initial pose proposal, which then were refined iteratively. RePOSE \cite{iwase2021repose} introduced the intermediate representation with a neural network to calculate the relative transformation. In addition to new representation, RNNPose \cite{xu2022rnnpose} exploited the recurrent neural network in the iteration process. \citeauthor{lipson2022coupled} \cite{lipson2022coupled} iteratively refined pose and correspondence in a tightly coupled manner to be more robust to outliers. DenseFusion \cite{wang2019densefusion} extracted the global features in the point cloud, and then the object pose was refined as iteratively reducing the residual between these global features. Although these methods generally improved the pose accuracy, they still suffer from occlusion between objects as the single-image approach.

\subsection{Multiple-View-Based Pose Refinement}

Researchers have utilized sequential multi-view information to achieve more accurate 6D poses and deal with the inherent problem of single-view-based approaches. se(3)-TrackNet \cite{wen2020se} estimated the object poses by performing pose residual estimation between the current observation and the synthetic rendered image for 6D object tracking. \citeauthor{deng2020self} \cite{deng2020self} enhanced their previous method, PoseRBPF, in a self-supervised manner along with their additional data sequences. These methods did not guarantee global consistency as a tracking approach that utilizes the information from the previous frame. MoreFusion \cite{wada2020morefusion} generated the occupancy map using sequential RGBD views, and then performed joint optimization to estimate collision-free multi-object poses.
CosyPose \cite{labbe2020cosypose} removed the non-consistent object pose proposal through inlier checking. Then, this 6D object pose was optimized through consistent matching between the initial pose proposals estimated in the multi-view input images. \citeauthor{shugurov2021multi} \cite{shugurov2021multi} exploited the differentiable rendering to make geometrical constraints using the relative camera poses. BundleTrack \cite{wen2021bundletrack} estimated the object pose proposal using keypoint detection, then performed pose-graph optimization for selected keyframes. \citeauthor{Merrill2022CVPR} \cite{Merrill2022CVPR} proposed an object-level \ac{SLAM} based on keypoints extracted from the network. These methods required prior information about shape attributes and may encounter the best-view selection problem. Unlike these methods, our work maintains global consistency through graph optimization with all available data, and does not need prior information about object shape.

\subsection{Object Pose Uncertainty Estimation}


To integrate deep-learning methods with other modules in robotic tasks, uncertainty awareness for a deep-learning module becomes essential. For the object pose uncertainty, PoseRBPF sampled the particles through bounding-box variation for translation distribution. The orientation distribution was obtained by comparing the discretized rotation codebook and the sampled particles. The estimated distribution was propagated with a particle filter. \citeauthor{Shi2021fast} \cite{Shi2021fast} proposed the ensemble-based uncertainty quantification. In \cite{okorn2020learning}, the authors regressed the object orientation and its uncertainty based on the Bingham distribution. \citeauthor{deng2022deep} \cite{deng2022deep} regressed Bingham mixture models through multi-hypothesis prediction. \citeauthor{manhardt2019explaining} \cite{manhardt2019explaining} estimated specific pose distribution from multiple object pose hypotheses. \citeauthor{Merrill2022CVPR} \cite{Merrill2022CVPR} predicted the Gaussian covariance of object keypoints to utilize to keypoints-based \ac{SLAM}. Instead, we regress the uncertainties of reconstructed primitive images, and then combine these uncertainties for the initial pose proposal and multi-object pose optimization.
\section{Method}
\label{sec:method}

\subsection{Notation}

We define the notation to describe our proposed method. The transformation from coordinate system $A$ to coordinate system $B$ is denoted as $^{A}_{B}T$, which comprises of rotation $\mathbf{R}_{AB} \in SO(3)$ and translation $\mathbf{t}_{AB} \in \mathbb{R}^3$. ${W}$ represents the world coordinate. The subscript $(\cdot)_{i}$ indicates the specific index of the corresponding variable. The subscript $(\cdot)_{0:n}$ indicates the group comprising the initial value to the $n^{\text{th}}$ value.



\subsection{Overview}

Given a manipulator equipped with an RGB-D camera that moves over a set of objects in the workspace, the robotic system collects RGB-D images captured from various viewpoints and the associated joint encoder angles of the robot manipulator. Taking all of this sequential data as input, the proposed method aims to optimize the multi-object pose by fusing the initial object poses from the RGB camera and other heterogeneous sensor data. As shown in \figref{fig:overview}, the proposed method is two-staged. In the first stage, we use RGB images to estimate simultaneously generic primitive uncertainty and 6D object pose $^{C}_{O}T$ expressed in terms of its relative transform between camera coordinate frame $C$ and object coordinate frame $O$. In the second stage, we refine object poses in an object \ac{SLAM} manner using camera poses via factor graph optimization.


\subsection{6D Object Pose Initialization}

\subsubsection{Object Segmentation}


First, we performe object segmentation. We construct a segmentation network adopting the encoder-decoder architecture based on ResNet34. \cite{he2016deep}. Overall, the network generates the $N+1$ dimension segmentation map, where $N$ indicates the number of objects.

\subsubsection{Rotational Primitive Axis Reconstruction}

Two major challenges in object pose estimation are occlusion and symmetry. To overcome these challenges, we leverage ideas from our previous work \cite{jeon2020prima6d} on primitive reconstruction for objects followed by keypoint extraction from these primitives. By doing so, we can reliably estimate the pose from even small and texture-less objects.

Similarly, as in \cite{jeon2020prima6d}, we cropp the input RGB image and concentrate on the target object using the segmentation map. We then transform the cropped RGB image into the primitive image using the \ac{AE}-based primitive reconstruction network. Differing from \cite{jeon2020prima6d}, our network separately predicts the three rotation axis primitive images per target object by appending three additional layers to the decoder. Each primitive image represents a respective rotation axis of the 6D object pose and each will be used to predict the uncertainty in the next stage.

To train the primitive reconstruction network, we use a primitive reconstruction loss $\mathbf{L}_{R}$ \cite{jeon2020prima6d} for each a rotation axis primitive image, $\mathcal{P}_{x}$, $\mathcal{P}_{y}$, and $\mathcal{P}_{z}$.

\small
\begin{equation}
  	\label{eq:primitive}
  	\mathcal{L_{P}} = \mathbf{L}_{R}(\mathcal{P}_{x}) + \mathbf{L}_{R}(\mathcal{P}_{y}) + \mathbf{L}_{R}(\mathcal{P}_{z})
\end{equation}
\normalsize

To improve the quality of the primitive reconstruction, we adopt the adversarial loss where the generator and the discriminator are each focused on minimizing $\mathcal{L_{G}}=\mathbb{E}_{\hat{X}}[\log \mathcal{D}(\hat{X})]$ and maximizing $\mathcal{L_{D}}=\mathbb{E}_{X}[\log (1-\mathcal{D}(\mathcal{G}(X)))]$ over target primitive image $X$ and estimated primitive image $\hat{X}$.


\subsubsection{Keypoints and Uncertainty Estimation}

Given the reconstructed primitive images, we extract keypoints and the associated uncertainty. Even an asymmetric target object might reveal ambiguity depending on the viewpoint yielding obscurity in the reconstructed primitive image as shown from two samples in \figref{fig:intro}. We exploit this degeneracy when simultaneously regressing the keypoints of the rotation axis primitive images and their associated uncertainties. Our intention is to estimate the uncertainty by measuring the quality of the reconstructed rotation axis primitive image.

To train the keypoints and uncertainty simultaneously, we assume the uncertainty as a zero-mean normal distribution $\mathcal{N}(0, \Sigma)$ and estimate $\Sigma$ by minimizing the negative log-likelihood of distribution over the keypoint errors $e_{k}=| x_{k}-\hat{x}_{k} |$ over $n$ keypoints in the training set.


%
\small
\begin{eqnarray}
	\label{eq:keypoint1}
    \mathbf{L}(e) &=& \underset{\Sigma_{i:n}}{\mathrm{argmin}} \sum_{i=1}^{n} -\log p(e_{i}|\Sigma_{i}) \\
	\label{eq:keypoint2}
	&=& \underset{\Sigma_{i:n}}{\mathrm{argmin}} \sum_{i=1}^{n} \log | \Sigma_{i} | + e^{\top}_{i}\Sigma_{i}^{-1}e_{i}
\end{eqnarray}
\normalsize

The keypoints are independent, having a positive definite covariance matrix. We assume the diagonal terms of covariance matrix $\Sigma$ would have the same value by defining the covariance matrix as $\Sigma = \operatorname{diag} (\sigma) = \operatorname{diag} (\mathbf{\exp(\xi)})$. Using \cite{liu2018deep}, \eqref{eq:keypoint2} can be re-written as:

\small
\begin{equation}
	\label{eq:keypoint3}
	\mathbf{L}(e) = \underset{\sigma}{\mathrm{argmin}} \sum_{i=1}^{n} \text{tr}(\log \Sigma) +  e^{\top}_{i}\Sigma^{-1}e_{i}
\end{equation}
\normalsize

As shown in \figref{fig:overview}, each rotation axis of the primitive image contains 14 keypoints. each axis has colored (red, blue, or green) and white regions. Although we estimate the axis separately at the uncertainty estimation stage, we will conbime the three axes later, sharing the white region. The keypoints consist of one centered point and four cornered points in the colored region and as well as centered point and eight cornered points in the white region. One centered point in the white region coincides with the center of the object. We build the keypoint and uncertainty estimation network based on ResNet18 \cite{he2016deep}. To train this network, we use \eqref{eq:keypoint3} per rotation axis primitive image, resulting in \eqref{eq:keypoint4}, allowing us to train the keypoints in a supervised manner and the correlated uncertainty in an unsupervised manner, respectively. Consequently, we obtaine three sets $\{ \mathcal{K}_{x}, \mathcal{K}_{y}, \mathcal{K}_{z} \}$ of 14 keypoints and three uncertainties $\boldsymbol{\sigma} = \{ \sigma_{x}, \sigma_{y}, \sigma_{z} \}$ per input image.

\small
\begin{equation}
	\label{eq:keypoint4}
	\mathcal{L_{K}} = \mathbf{L}(e_{\mathcal{K}_{x}}) + \mathbf{L}(e_{\mathcal{K}_{y}}) + \mathbf{L}(e_{\mathcal{K}_{z}})
\end{equation}
\normalsize
, while $e_{\mathcal{K}}$ is the error measured over a set $\mathcal{K}$.

\begin{figure}[t]
	\begin{algorithm}[H]
		\algrenewcommand\alglinenumber[1]{#1:}
		\caption{Uncertainty-Based Keypoints Selection}
		\label{alg:keypoint}
		\hspace*{\algorithmicindent} \textbf{Input:} $i \in \{x, y, z\}$ \\
		\hspace*{\algorithmicindent} Estimated Keypoints: $\mathcal{K}_{i}=\{ \mathcal{K}_{i1}, \cdots, \mathcal{K}_{i14} \}$, \\
		\hspace*{\algorithmicindent} Estimated Uncertainties: $\sigma_{i}$\\
		\hspace*{\algorithmicindent} Threshold: $\sigma_{o}$\\
		\hspace*{\algorithmicindent} \textbf{Output:} Merged Keypoints: $\mathcal{K}_{m}$
		\begin{algorithmic}[1]
			\State $N=0, x_{1:9}=0$
			\For {$i \ \textbf{in} \ \{x, y, z\}$}
			\If {$\sigma_{i} > \sigma_{o}$}
			\State continue
			\Else
			\State $N=N+1$
			\For {$j=1:9$}
			\State $x_{j}=\frac{N-1}{N}x_{j}+\frac{1}{N}\mathcal{K}_{ij}$
			\EndFor
			\State $\mathcal{K}_{m}.append(\mathcal{K}_{i10:i14})$
			\EndIf
			\EndFor
			\If {size($\mathcal{K}_{m}$) == 0}
			\State \textbf{Return} False 
			\EndIf
			\State $\mathcal{K}_{m}.append(x_{1:9})$
			\State \textbf{Return} $\mathcal{K}_{m}$
		\end{algorithmic}
	\end{algorithm}
\end{figure}

\subsubsection{Uncertainty-Based Keypoints Selection}
\label{sec:k_sel}

As the final step in the pose initialization, we reject ambiguous primitive axes using the predicted uncertainty. The rotation axis in the primitive image may not be impeccably reconstructed due to occlusion, and symmetry. In such degenerate cases, inferring the 6D object pose using all keypoints rather deteriorates the accuracy, therefore, we need to reject the ambiguous rotation axis based on the estimated uncertainty. Among three rotation axis primitive images $\{ \mathcal{P}_{x}, \mathcal{P}_{y}, \mathcal{P}_{z} \}$ of an object, we examine the uncertainty of each primitive image to check validity (smaller than the threshold $\sigma_{o}$).

We discard the invalid axis and merge the keypoints only from the valid rotation axes. \figref{fig:keypoints} shows exemplar primitives from objects. Keypoints from the colored region of the primitive can be immediately confirmed as the final keypoints, whereas the keypoints in the shared white region are averaged over the valid primitives. By averaging, we can ensure a more stable object center. The uncertainty of each object pose factor is the norm of the uncertainties for valid rotation axis primitive images. The algorithm for completing keypoints is \algoref{alg:keypoint}.

Last, we transform the keypoints back to the whole image plane using the cropped bounding box information, then we calculate the rotation and translation of the object through \ac{PnP} using the confirmed keypoints. As a further enhancement before adding the estimated object pose to the factor graph, for the cases where the norm $\left\| {\boldsymbol{\sigma}} \right\|$ of valid uncertainties is lower than a threshold $\sigma_{icp}$, we perform \ac{ICP} refinement using depth measurement.



\subsection{Object Pose Refinement via Graph SLAM}

\begin{figure}[!t]
	\centering  
	\includegraphics[width=0.9\columnwidth]{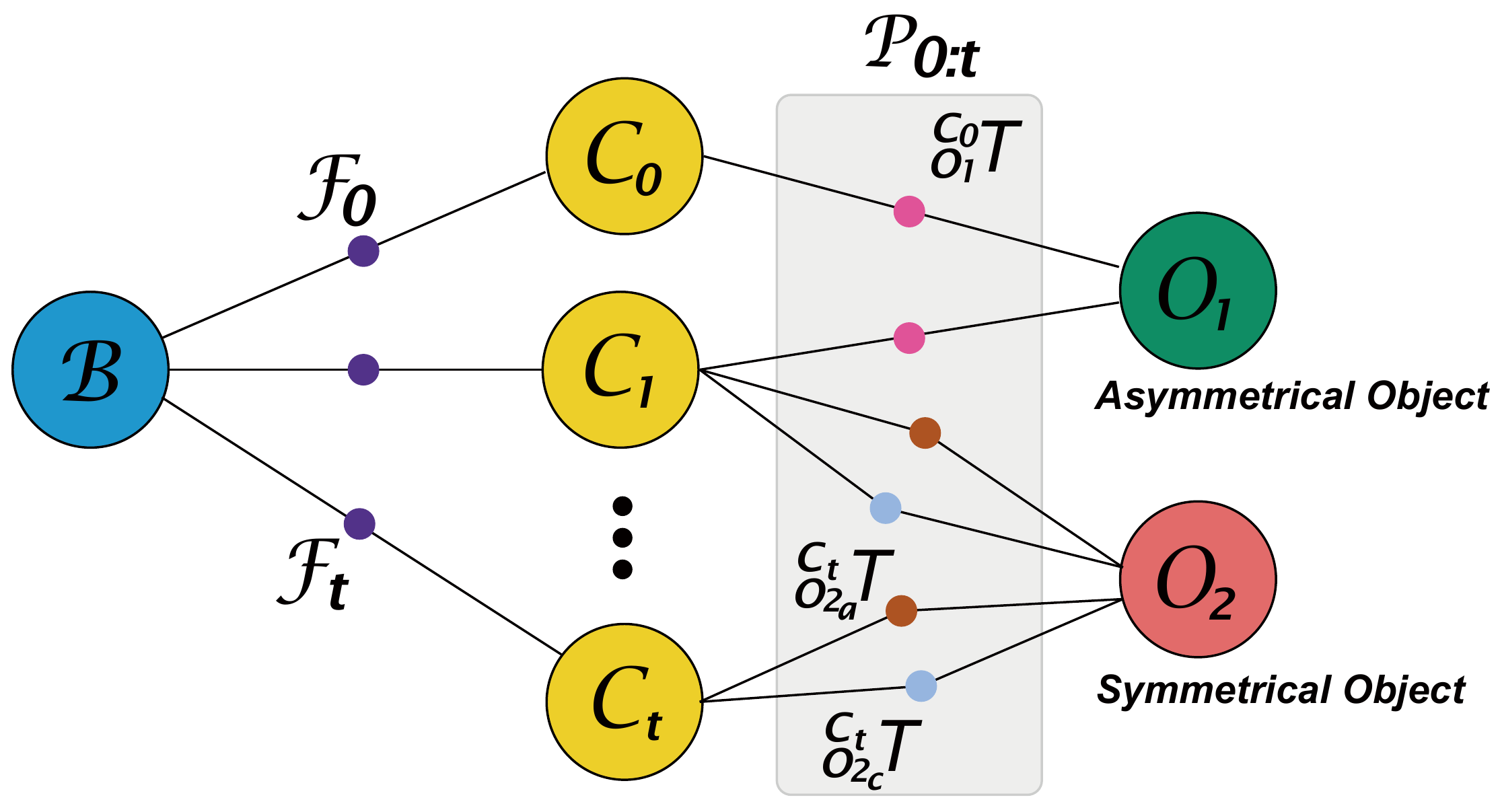}
	\caption{The factor graph illustration. The colored circles depict the state variables. Two independent measurements are the factors represented by lines that constrain the state at time $t$. On every occurrence of the 6D object pose estimation, two factors ($\mathcal{F}$, $\mathcal{P}$) are added to the factor graph. The object pose factor $\mathcal{P}$ is added in two ways depending on the type of object.}
	\label{fig:graph} 
  \end{figure}

The computed initial relative pose between an object and the camera can be viewed as a factor in the object graph SLAM. We assume that the base of the robot manipulator, and the set of objects are stationary and that the relative transformation between the end-effector of the robot manipulator and the camera is given. Based on these assumptions, we constitute the graph \ac{SLAM} by incorporating three states and two measurement types (\figref{fig:graph}).

The state variables consist of the camera pose attached to the end-effector of the robot manipulator $\mathcal{C}_{0:t} \triangleq \{ {}^{W}_{C_{i}}T \in SE(3) :\forall i \in [0, t]\}$ and the $k$ number of object pose $\mathcal{O}_{1:k} \triangleq \{ {}^{W}_{O_{i}}T \in SE(3) :\forall i \in [1, k] \}$ expressed in the base pose of robot manipulator $\mathcal{B} \triangleq \{ {}^{W}_{B}T \in SE(3) \}$ upto current time $t$.

\small
\begin{equation}
  \label{eq:state}
  \mathcal{X}_{t} = \{\mathcal{B}, \ \mathcal{C}_{0:t}, \mathcal{O}_{1:k}\}
\end{equation}
\normalsize

The two types of measurement are comprised of camera pose measurement $\mathcal{F}_{t}\triangleq \{ {}^{B}_{C_{t}}T \in SE(3) \}$ and object pose measurement $\mathcal{P}_{t}\triangleq \{ {}^{C_{t}}_{O_{i}}T \in SE(3) : \forall i \in [1, k] \}$ at time $t$.

\small
\begin{equation}
	\label{eq:measurement}
    \mathcal{Z}_{t} = \{ \mathcal{F}_{t}, \ \mathcal{P}_{t} \}
\end{equation}
\normalsize

Since the measurement $\mathcal{Z}_{t}$ only depends on the state $\mathcal{X}_{t}$, the posterior probability of our state estimation problem can be written as $P(\mathcal{X}_{t}|\mathcal{Z}_{t})$.


%
\small
\begin{eqnarray}
	\label{eq:map}
	&&p(\mathcal{X}_{t}|\mathcal{Z}_{t}) \varpropto p(\mathcal{F}_{t}, \mathcal{P}_{t}|\mathcal{B}, \mathcal{C}_{0:t}, \mathcal{O}_{1:k}) 	\\
	\label{eq:map1}
	&&= \prod_{i=0}^{t}p(\mathcal{F}_{i}|\mathcal{B}, \mathcal{C}_{i}) \prod_{i=0}^{t}\prod_{j=1}^{k}p(\mathcal{P}_{i}|\mathcal{C}_{i}, \mathcal{O}_{j})
\end{eqnarray}
\normalsize

\subsubsection{Camera Pose Factor}
\label{sec:fk_factor}

The first term of \eqref{eq:map1} presents the relative transform between the base $\mathcal{B}$ and the attached camera $\mathcal{C}_{t}$, which is obtained from the robot forward kinematics using the joint angles. We utilize a standard residual error model for the measurement $\mathcal{F}_{t}$ of the camera pose.


%
\small
\begin{eqnarray}
	\label{eq:resudual_ee}
	\Delta \mathbf{t}_{\mathcal{B}\mathcal{C}_{t}} &=& \mathbf{R}_{\mathcal{B}}^{\top}(\mathbf{t}_{\mathcal{B}}-\mathbf{t}_{\mathcal{C}_{t}}) + \delta \mathbf{t}_{\mathcal{B}\mathcal{C}_{t}} \\
	\Delta \mathbf{R}_{\mathcal{B}\mathcal{C}_{t}} &=& \mathbf{R}_{\mathcal{B}}^{\top}\mathbf{R}_{\mathcal{C}_{t}} \mathrm{Exp}(\delta \phi_{\mathcal{B}\mathcal{C}_{t}}) \\
	\left\| \mathbf{r}_{\mathcal{B}\mathcal{C}_{t}} \right\|^{2} &=& \left\| \delta\mathbf{t}_{\mathcal{B}\mathcal{C}_{t}} \right\|^{2}_{\Sigma_{\mathcal{C}_{t}}} + \left\| \delta\phi_{\mathcal{B}\mathcal{C}_{t}} \right\|^{2}_{\Sigma_{\mathcal{C}_{t}}}
\end{eqnarray}
\normalsize

\begin{figure}[!t]
	\centering
	\includegraphics[width=0.9\columnwidth]{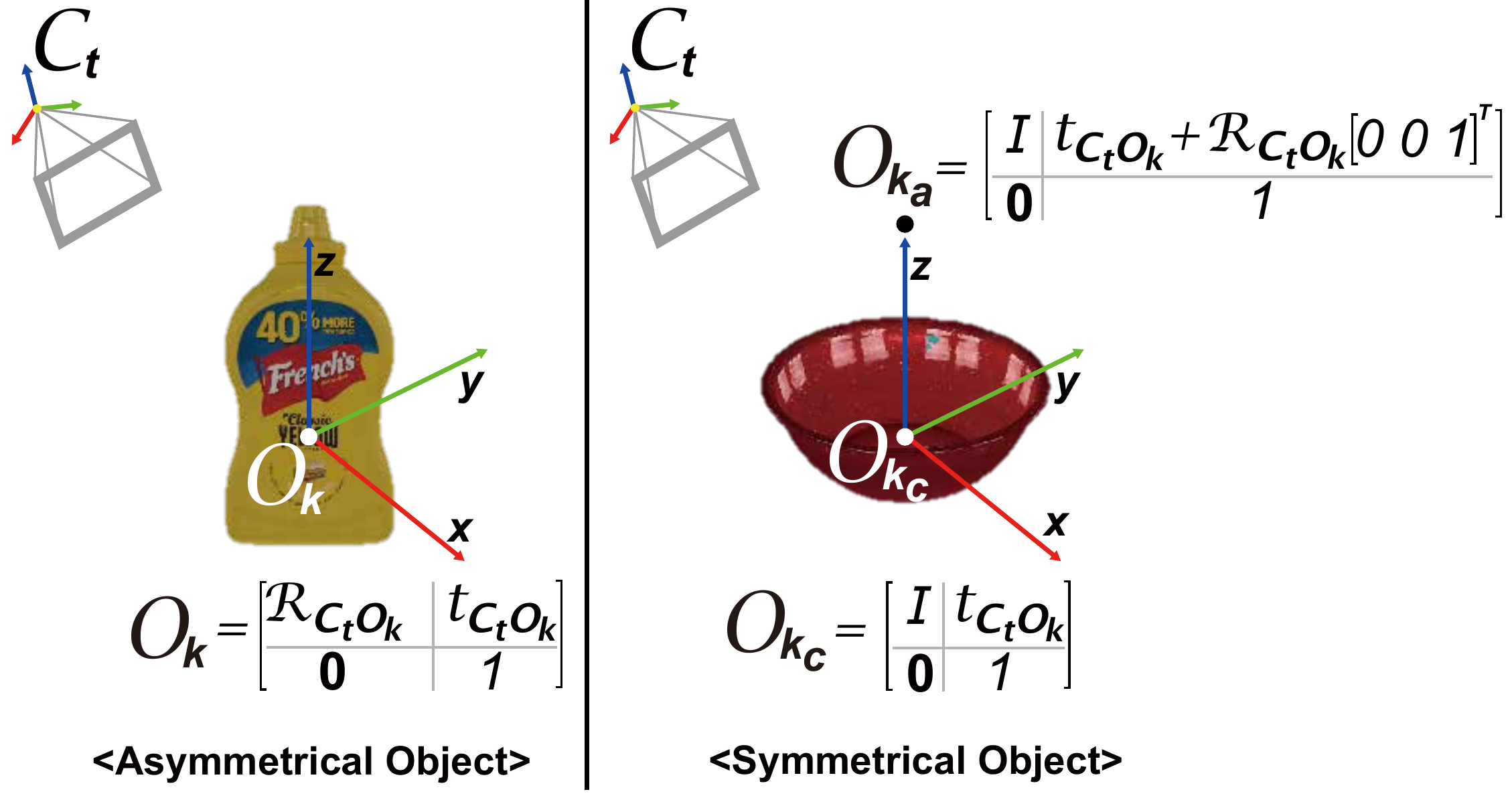}
	\caption{The constraint points definition for the object pose factor.}
	\label{fig:factor}
\end{figure}

\subsubsection{Object Pose Factor}

The second term of \eqref{eq:map1}, the object pose factor, describes the relative transform between the camera $\mathcal{C}_{t}$ and target object $\mathcal{O}_{k}$. We obtain two types of object states from PrimA6D++ as shown below.

\textbf{\textit{(i)} Asymmetric case:} In the asymmetric cases, we can simply express the residual error in \secref{sec:fk_factor}.


%
\small
\begin{eqnarray}
	\left\| \mathbf{r}_{\mathcal{C}_{t}\mathcal{O}_{k}} \right\|^{2} &=& \left\| \delta\mathbf{t}_{\mathcal{C}_{t}\mathcal{O}_{k}} \right\|^{2}_{\Sigma_{\mathcal{C}_{t}}} + \left\| \delta\phi_{\mathcal{C}_{t}\mathcal{O}_{k}} \right\|^{2}_{\Sigma_{\mathcal{C}_{t}}}
\end{eqnarray}
\normalsize

\textbf{\textit{(ii)} Symmetric case:} To cope with the potential ambiguity in the symmetric cases, we define a factor utilizing the dominant rotation axis of the object. The dominant axis of the object is predefined as the main symmetrical axis of the object based on the 3D CAD model. We begin by defining the dominant axis using two points on the dominant vector, where the object center $\mathcal{O}_{k_{c}}$ and an arbitrary point $\mathcal{O}_{k_{a}}$ lie on the dominant axis (\figref{fig:factor}).


%
\small
\begin{eqnarray*}
	\mathbf{t}_{\mathcal{C}_{t} \mathcal{O}_{k_{c}}} &=& \mathbf{t}_{\mathcal{C}_{t}\mathcal{O}_{k}}, \  \mathbf{R}_{\mathcal{C}_{t} \mathcal{O}_{k_{c}}}=\mathbf{R}_{\mathcal{C}_{t} \mathcal{O}_{k_{a}}}=\mathbf{I} \\
	\mathbf{t}_{\mathcal{C}_{t} \mathcal{O}_{k_{a}}} &=& \mathbf{t}_{\mathcal{C}_{t}\mathcal{O}_{k}} + \mathbf{R}_{\mathcal{C}_{t}\mathcal{O}_{k}} \left [0 \ 0 \ 1 \right ]^{\top}
\end{eqnarray*}
\normalsize

The residual error is then computed using the transform between these two points $\mathcal{O}_{k_{c}}$, $\mathcal{O}_{k_{a}}$ and the camera $\mathcal{C}_{t}$.


%
\small
\begin{eqnarray}
	\left\| \mathbf{r}_{\mathcal{C}_{t}\mathcal{O}_{k_{c}}} \right\|^{2} &=& \left\| \delta\mathbf{t}_{\mathcal{C}_{t}\mathcal{O}_{k_{c}}} \right\|^{2}_{\Sigma_{\mathcal{C}_{t}}} + \left\| \delta\phi_{\mathcal{C}_{t}\mathcal{O}_{k_{c}}} \right\|^{2}_{\Sigma_{\mathcal{C}_{t}}} \\
	\left\| \mathbf{r}_{\mathcal{C}_{t}\mathcal{O}_{k_{a}}} \right\|^{2} &=& \left\| \delta\mathbf{t}_{\mathcal{C}_{t}\mathcal{O}_{k_{a}}} \right\|^{2}_{\Sigma_{\mathcal{C}_{t}}} + \left\| \delta\phi_{\mathcal{C}_{t}\mathcal{O}_{k_{a}}} \right\|^{2}_{\Sigma_{\mathcal{C}_{t}}}
\end{eqnarray}
\normalsize

The initial covariance of the object pose is defined by $\Sigma_{\mathcal{C}_{t}} =  \operatorname{diag} (\left \| \boldsymbol{\sigma}_{t} \right \|)$.


\begin{figure*}[!t]
	\centering
	\begin{minipage}{0.7\textwidth} \centering
		\subfigure[Uncertainty and keypoint error]{%
		  \includegraphics[width=1.0\columnwidth]{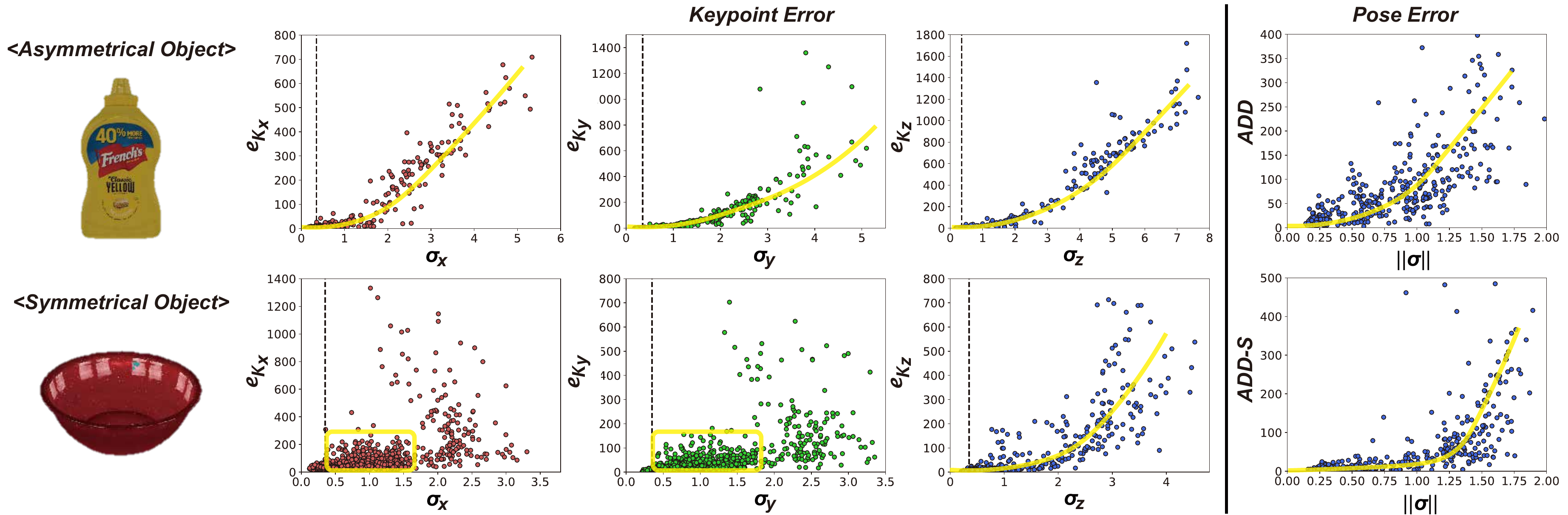}
		  \label{fig:scatter_plot}
		}
  \end{minipage}
  \begin{minipage}{0.27\textwidth} \centering
		\begin{minipage}{0.7\textwidth} \centering
			\subfigure[Threshold selection]{%
        \includegraphics[width=1.0\textwidth]{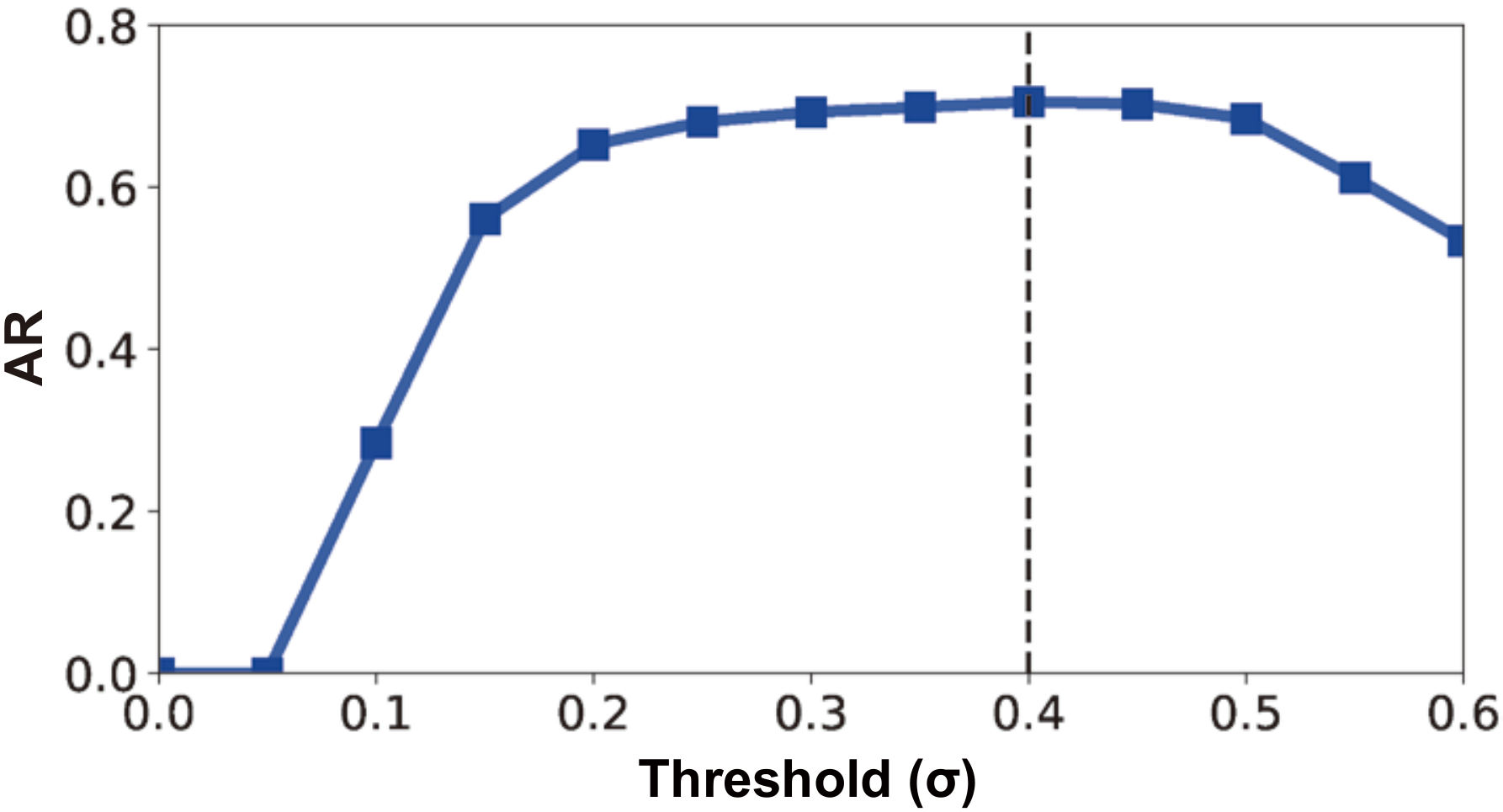}
				\label{fig:ablation}
			}
    \end{minipage}\\
		\begin{minipage}{0.7\textwidth}	
			\centering
			\resizebox{\textwidth}{!}{
				\begin{tabular}{ccc}
					\hline
					\multicolumn{1}{c|}{}              & Full Axis & Valid Axis \\ \hline \hline
					\multicolumn{1}{c|}{$\mathrm{AR_{vsd}}$} & 0.499     & 0.587      \\ \hline
					\multicolumn{1}{c|}{$\mathrm{AR_{mssd}}$} & 0.536     & 0.636      \\ \hline
					\multicolumn{1}{c|}{$\mathrm{AR_{mspd}}$} & 0.708     & 0.897      \\ \hline
					\multicolumn{1}{c|}{$\mathrm{AR}$} & 0.581     & 0.706      \\ \hline
				\end{tabular}	
			}
			\captionof{table}{Performance variance}
		\end{minipage}
  \end{minipage}
	\caption{\subref{fig:scatter_plot} The scatter plots represent the relation between the estimated uncertainties and keypoint errors and the relation between the norm of estimated uncertainties and ADD(-S) error. The first row illustrates the \texttt{mustard bottle} of the YCB-Video dataset as the typical asymmetrical object. The second row illustrates the \texttt{bowl} of the YCB-Video dataset as the representative symmetrical object. The black vertical dashed line is a threshold $\sigma_{o} = 0.4$. It can be observed that higher uncertainties lead to higher keypoint errors and object pose errors. \subref{fig:ablation} The graph depicts the performance variation with the threshold $\sigma_{o}$. The black vertical dashed line is a threshold $\sigma_{o} = 0.4$ used in our system. The table represents the performance variance according to uncertainty-based keypoints selection.}
\end{figure*}

\subsubsection{Pose Graph Optimization}

In our approach, we optimize the multi-object poses using pose graph optimization. Assuming that measurement noises are zero-mean normal distributions $\mathcal{N}(0, \Sigma)$, the \ac{MAP} estimate of state $\mathcal{X}_{t}$ is obtained by minimizing the negative log of probability, which is the same as minimizing the sum of squared residual errors.


%
\small
\begin{eqnarray}
	\hat{\mathcal{X}_{t}} &=& - \underset{\mathcal{X}_{t}}{\mathrm{argmin}} \ \log p(\mathcal{X}_{t}|\mathcal{Z}_{t}) \\
	&=& \underset{\mathcal{X}_{t}}{\mathrm{argmin}} \sum_{i=0}^{t} \left\| \mathbf{r}_{\mathcal{B}\mathcal{C}_{i}} \right\|^{2}_{\Sigma_{\mathcal{C}_{i}}} + \alpha \left ( \sum_{i=0}^{t} \sum_{j=1}^{k} \left\| \mathbf{r}_{\mathcal{C}_{i}\mathcal{O}_{j}} \right\|^{2}_{\Sigma_{\mathcal{C}_{i}}} \right ) + \nonumber \\
	&&(1 - \alpha ) \left ( \sum_{i=0}^{t} \sum_{j=1}^{k} \left\| \mathbf{r}_{\mathcal{C}_{i}\mathcal{O}_{j_{c}}} \right\|^{2}_{\Sigma_{\mathcal{C}_{i}}} + \sum_{i=0}^{t} \sum_{j=1}^{k} \left\| \mathbf{r}_{\mathcal{C}_{i}\mathcal{O}_{j_{a}}} \right\|^{2}_{\Sigma_{\mathcal{C}_{i}}} \right ) \nonumber \\
	\nonumber \\
	\alpha &=&
	\begin{cases}
		1, & \text{if } \mathcal{O}_{j} \text{ is asymmetrical case} \\
		0, & \text{otherwise}
	\end{cases}
\end{eqnarray}
\normalsize

\section{Experiment}
\label{sec:experiment}

In this section, we mainly evaluate (\textit{i}) How much the estimated uncertainties associate the 6D object pose error, (\textit{ii}) the accuracy of the estimated object pose, and (\textit{iii}) the effect of the refinement via factor graph optimization.

\subsection{Dataset and Implementation Detail}

In total, two datasets were used for the evaluation: T-LESS dataset \cite{hodan2017tless} and YCB-Video \cite{xiang2018posecnn}. The two datasets provide the relative transform between the world and camera coordinate per scene. The YCB-Video dataset is composed of 21 high-quality 3D models, offering 92 annotated video sequences. These sequences include various lighting conditions, noise in the capture, and occlusion. The T-LESS dataset is composed of 30 3D models and offers 20 annotated video sequences. This dataset is challenging given that most objects are symmetrical and texture-less.

For training, we synthesized 100K images per the object together with their associated three rotation axis primitive images. The synthetic images were further augmented using the ``imgaug'' library \cite{imgaug}. We also utilized 50K photorealistic synthetic images from \cite{hodavn2019photorealistic} and other available real image training set from public datasets. The proposed network was implemented using Pytorch and was trained with NVIDIA RTX 2080Ti. All of the networks were trained using Adam optimizer with $\beta_{1} = 0.9$, $\beta_{2} = 0.999$, and a learning rate of $10^{-4}$. The segmentation network was trained for 50 epochs with a batch size of 10. Input images with a resolution of 640$\times$480 were used, and domain randomization was used as the data augmentation. Other networks were trained together for 220 epochs with a batch size of 100. Here, we used input images with a resolution of 64$\times$64 for training. For single-view inference, we set a threshold $\sigma_{o}=0.4$. In the optimization phase, we built and solved the factor graph using gtsam \cite{dellaert2012factor} with the threshold $\sigma_{icp}=0.4$.


\subsection{Evaluation Metrics}

\textbf{(i) Average distance metric:} Similar to \cite{xiang2018posecnn}, average distance metric ADD(-S) is selected for evaluation. ADD(-S) computes the distance of the transformed 3D model points between the ground truth pose and predicted pose. We also compute the area under the curve varying the distance threshold to evaluate performance of pose estimation.

\textbf{(ii) Standard visual surface discrepancy:} We measure the distance between the estimated and ground truth of visible surface on object depth. As in \cite{sundermeyer2018implicit}, we consider the sample as a true positive case if the error is less than 0.3 on the tolerance $\tau=$\unit{20}{mm} and visibility $v > 0.1$.

\textbf{(ii) BOP challenge metrics:} We use the metrics utilized in the BOP challenge \cite{hodan2018bop}. It evaluates the average recall of errors measured by Visible Surface discerpancy (VSD), Maximum Symmetry-Aware Surface Distance (MSSD), and Maximum Symmetry-Aware Projection Distance (MSPD). The overall performance is calculated as average recall $\mathrm{AR} = (\mathrm{AR_{vsd}} + \mathrm{AR_{mssd}} + \mathrm{AR_{mspd}}) / 3$.

\subsection{Evaluation of Uncertainty}

We evaluate the correlation between the estimated uncertainties and the associated 6D object pose. We further discuss the uncertainty variation under object symmetricity and occlusions and the effect of ambiguity rejection.

\begin{table*}[!h]
    \centering
    \begin{adjustbox}{width=1.0\linewidth}
    {
        \begin{tabular}{c|ccccc|ccccc}
            \hline
                                     & \multicolumn{5}{c|}{Single-View}                                                                                                                                                                                     & \multicolumn{4}{c}{Multi-View}                                                                                                                                                                                                                                               \\ \hline
            \multirow{2}{*}{methods} & \multirow{2}{*}{GDR-Net \cite{wang2021gdr}} & \multirow{2}{*}{CosyPose \cite{labbe2020cosypose}} & \multirow{2}{*}{PrimA6D++} & \multirow{2}{*}{MoreFusion \cite{wada2020morefusion}} & \multirow{2}{*}{\begin{tabular}[c]{@{}c@{}}FFB6D \cite{He_2021_CVPR}\\ w/ ICP\end{tabular}} & \multirow{2}{*}{SUO-SLAM \cite{Merrill2022CVPR}} & \multirow{2}{*}{\begin{tabular}[c]{@{}c@{}}CosyPose \cite{labbe2020cosypose}\\ w/ 8-View\end{tabular}} & \multirow{2}{*}{\begin{tabular}[c]{@{}c@{}}PrimA6D++\\ w/ Graph Opt\end{tabular}}& \multirow{2}{*}{\begin{tabular}[c]{@{}c@{}}PoseRBPF \cite{deng2021poserbpf}\\ w/ SDF\end{tabular}} & \multirow{2}{*}{\begin{tabular}[c]{@{}c@{}}PrimA6D++\\ w/ Graph Opt\end{tabular}} \\
                                     &                          &                           &                            &                           &                             &                                                                        &                           &                                                                              &                                                                                                                                                                  \\ \hline
            \#. models & N & 1 & N & 1 & N & 1 & 1 & N & N & N \\ \hline
            data type                & RGB                      & RGB                       & RGB                        & RGB-D                       & RGB-D                                                                  & RGB                       & RGB &                        RGB                                                 & RGB-D                                                                          & RGB-D                                                                            \\ \hline \hline
            ADD-(S)                  & 84.4                     & 84.5                      & 91.1                       & 91.0                        & 93.1                                                                   & 84.7                      & -               & 91.5                                                             & 87.5                                                                           & \textbf{94.4}                                                                             \\
            ADD-S                    & 91.6                     & 89.8                      & 94.3                       & 95.7                        & \textbf{97.0}                                                                     & 90.3                      & 93.4                     &94.8                                                    & 95.2                                                                           & 96.9                                                                             \\            
            \hline
            \end{tabular}
    }
    \end{adjustbox}
    \caption{The result of YCB-Video dataset on the ADD(-S) and ADD-S metrics. PrimA6D++ with graph optimization is divided into RGB and RGB-D depending on the presence of \ac{ICP} refinement. N refers to the total number of objects in the dataset. }
    \label{tab:ycb}
\end{table*}
\begin{table*}[!h]
    \centering
    \begin{adjustbox}{width=1.0\linewidth}
    {
        \begin{tabular}{c|ccccccc|ccc}
            \hline
                                     & \multicolumn{7}{c|}{Single View}                                                                                                                                                                                                                           & \multicolumn{2}{c}{Multi-View}                                                                                                                               \\ \hline
                                     &                          &                        &                          &                             &                            &                             &                                 &                           &                                                                                                                                                             \\
            \multirow{-2}{*}{method}  & \multirow{-2}{*}{CDPN \cite{li2019cdpn}} & \multirow{-2}{*}{WS-OPE \cite{li2022wsope}} & \multirow{-2}{*}{ZebraPose \cite{su2022zebrapose}} & \multirow{-2}{*}{CosyPose \cite{labbe2020cosypose}} & \multirow{-2}{*}{PrimA6D++} & \multirow{-2}{*}{{{ \cite{lipson2022coupled}}}} & \multirow{-2}{*}{SurfEmb \cite{haugaard2022surfemb}} & \multirow{-2}{*}{\begin{tabular}[c]{@{}c@{}}CosyPose \cite{labbe2020cosypose}\\ w/ 8-View\end{tabular}} & \multirow{-2}{*}{\begin{tabular}[c]{@{}c@{}}PrimA6D++ \\ w/ Graph Opt\end{tabular}} & \multirow{-2}{*}{\begin{tabular}[c]{@{}c@{}}PrimA6D++ \\ w/ Graph Opt\end{tabular}} \\ \hline
            \#. models & 1 & 1 & N & 1 & N & N & 1 & 1 & N & N \\ \hline
            data type                & RGB                    & RGB                      & RGB                         & RGB                        & RGB                         & RGB-D                                                  & RGB-D                     & RGB                                                                & RGB           & RGB-D                                                                        \\ \hline \hline
            $\mathrm{AR_{vsd}}$                   & 0.377                  & 0.476                    & 0.459                       & 0.669                      & 0.587                       & 0.760                                                   & \textbf{0.797}            & 0.773 & 0.692                                                                        & 0.752                                                                        \\
            $\mathrm{AR_{mssd}}$                  & 0.418                  & 0.587                    & 0.618                       & 0.695                      & 0.636                       & 0.773                                                  & 0.829                     & 0.836 & 0.920                                                                         & \textbf{0.953}                                                               \\ 
            $\mathrm{AR_{mspd}}$                  & 0.674                  & 0.652                    & 0.732                       & 0.821                      & 0.897              & 0.795                                                  & 0.859                     & 0.907  & 0.935                                                                       & \textbf{0.963}                                                                        \\ \hline
            $\mathrm{AR}$                       & 0.490                  & 0.572                    & 0.603                       & 0.728                      & 0.706                       & 0.776                                                  & 0.828                     & 0.839     & 0.849                                                                    & \textbf{0.889} \\ \hline                                                             
            \end{tabular}
    } 
    \end{adjustbox}   
    \caption{The result of the T-LESS dataset on the average recall metrics of three metrics used in the BOP challenge. PrimA6D++ with graph optimization is divided into RGB and RGB-D depending on the presence of \ac{ICP} refinement. N refers to the total number of objects in the dataset.}
    \label{tab:tless_bop}
\end{table*}

\begin{table}[!t]
    \centering
    \begin{adjustbox}{width=1.0\linewidth}
    {
        \begin{tabular}{c|cccc}
            \hline
            & \multicolumn{4}{c}{Multi-View} \\ \hline
            \multirow{2}{*}{method} & \multirow{2}{*}{SUO-SLAM \cite{Merrill2022CVPR}} & \multirow{2}{*}{\begin{tabular}[c]{@{}c@{}}CosyPose \cite{labbe2020cosypose}\\ w/ 8-View\end{tabular}} & \multirow{2}{*}{\begin{tabular}[c]{@{}c@{}}PoseRBPF \cite{deng2021poserbpf}\\ w/ SDF\end{tabular}} & \multirow{2}{*}{\begin{tabular}[c]{@{}c@{}}PrimA6D++ \\ w/ Graph Opt\end{tabular}} \\
            & & & & \\ \hline
            data type & RGB & RGB & RGB-D & RGB-D \\ \hline \hline
            $e_\mathrm{vsd}$ & 63.7& 71.6 & 82.6 & \textbf{87.8} \\ \hline
            \end{tabular}
    }
    \end{adjustbox}
    \caption{The result of the T-LESS dataset on the $e_\mathrm{vsd}$ metric.}
    \label{tab:tless_vsd}
\end{table}

\subsubsection{Uncertainty for Keypoint Error Measure}
\label{sec:exp_unc_key}

To validate capacity of uncertainty estimation quantitatively, we select two representative objects, one of each asymmetrical and symmetrical objects from the YCB-Video dataset. By performing image augmentation, we generate 1,000 image samples from each object having illumination and occlusion variation. Using these images, we examine whether the estimated uncertainty captures the true keypoint error.

The result is as shown in \figref{fig:scatter_plot}, where we present the relation between the estimated keypoints errors and uncertainties per axis. As for the asymmetrical object, as the keypoint errors of all axes increase, the uncertainties also increase. This means the estimated uncertainties reflect the tendency of the keypoint errors. For the case of the symmetrical object, the relation between the $z$-axis keypoint errors $e_{\mathcal{K}_{z}}$ and uncertainties $\sigma_{z}$ in the third column shows the monotonically increasing form. As for the other graphs, they hardly represent any correlation, since the primitive images of a symmetrical object are rarely reconstructed in its unique form except for the dominant axis. In particular, since multiple poses can correspond to a single state of the symmetrical object, we observe that in unstable regions, the keypoint errors fluctuate with increasing uncertainties. By doing so, the proposed approach estimates the 6D object pose in an identical way that can reject the ambiguous axes as the advantage of uncertainty estimation.

\begin{figure*}[!t]
	\centering
	\includegraphics[width=0.95\textwidth]{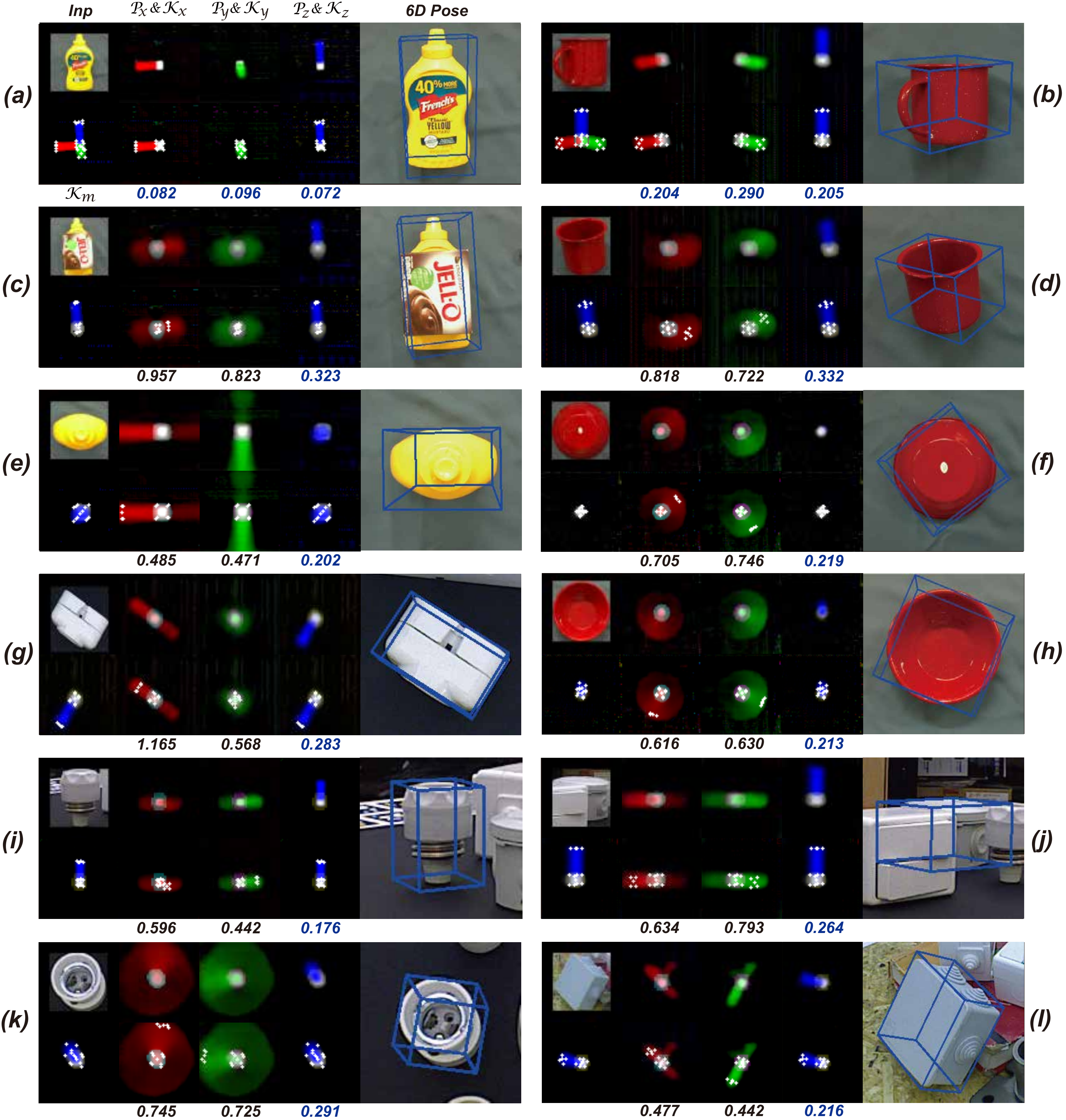}
  	\caption{The qualitative evaluation of ambiguity-aware 6D object pose estimation, PrimA6D++. The first column shows the cropped input image and merged keypoints obtained through uncertainty-based keypoints selection. Here, we set a threshold $\sigma_{o} = 0.4$. The second to fourth columns show the estimated rotation axis primitive images and their keypoints. The fifth column represents the 6D object pose estimated using merged keypoints. The below numbers indicate the estimated uncertainties corresponding to the primitive images.
      }
	\label{fig:keypoints}
\end{figure*}

\subsubsection{Threshold ($\sigma_{o}$) Selection}

Using these attributes, we can find the dominant axis and ambiguous axes of the symmetrical object based on the threshold $\sigma_{o} = 0.4$ and reject the ambiguous axes when inferring the 6D object pose. For this uncertainty-based selection, we have to set the proper threshold $\sigma_{o}$. In \figref{fig:ablation}, we describe the performance variation over the T-LESS dataset in terms of $\mathrm{AR}$ by increasing the threshold from 0 to 0.6 by 0.05 steps. When the threshold $\sigma_{o} = 0.4$, it achieves the highest performance, and further threshold increase causes a gradual performance drop.
In addition, by setting the threshold $\sigma_{o} = \infty$, the performance of using all axis primitives is shown in Table. \Romannum{1}. The case of inferring with all axis primitives cannot handle object ambiguity, thus the performance drops by 0.125 in terms of $\mathrm{AR}$. As a result, we select the threshold of $\sigma_{o} = 0.4$.
 
\begin{figure*}[!t]
	\centering
	\includegraphics[width=0.95\textwidth]{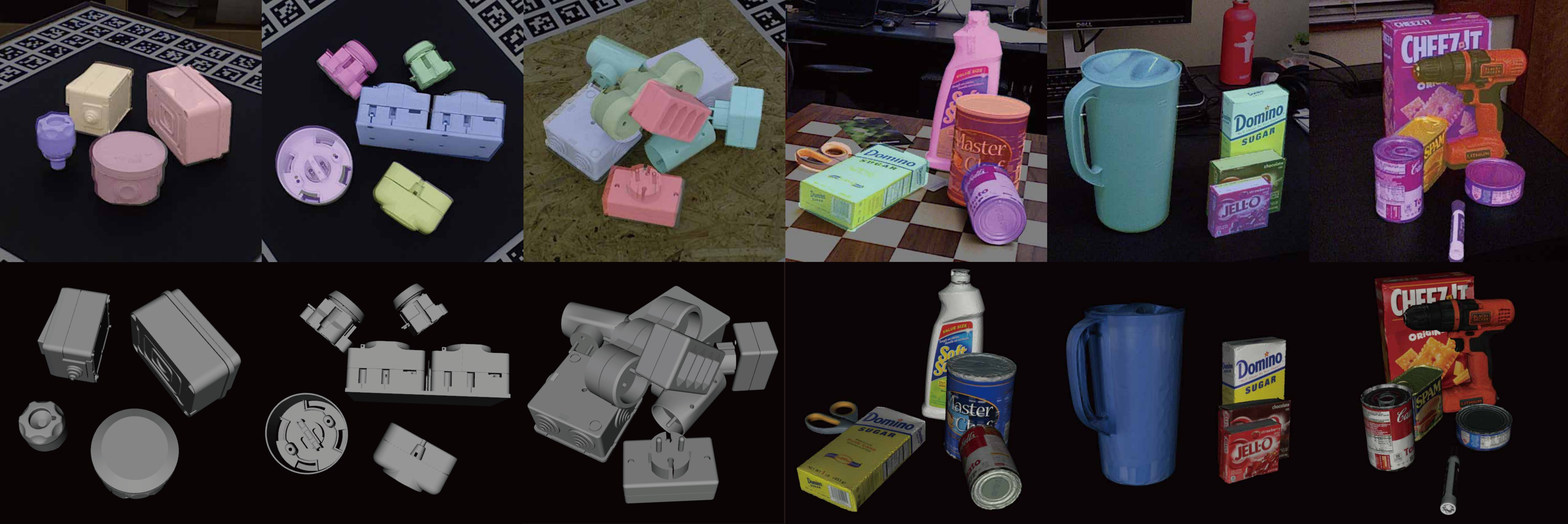}
  	\caption{The qualitative result of multi-object pose estimation. The first row represents the estimated poses and the second row is the reconstructed scenes using 3D object models on the T-LESS and YCB-Video dataset.
      }
	\label{fig:qualitative}
\end{figure*}

\subsubsection{Uncertainty for Ambiguity Awareness}
\label{sec:exp_unc_amb}

Next, we evaluate uncertainty as a measure of ambiguity awareness in a qualitative manner. We infer per-axis uncertainty over the T-LESS dataset and our own dataset using YCB objects. As mentioned earlier, even asymmetric objects may look symmetric depending on the viewpoint, and we validate whether this view-induced ambiguity is well detected from the reconstructed primitive images.

Exemplar cases of the object ambiguity awareness are presented in (\figref{fig:keypoints}). The merged keypoints are built using three valid axes for typical asymmetrical objects in (a) and (b). For the symmetrical object in (f), (h), (i), and (k) as well as the case of partial symmetricity in (g) and (l), the merged keypoints are built with the keypoints extracted from the valid primitives. Moreover, we can observe that the direction of the estimated dominant axis coincides with the objects as in (f) and (h). It is noteworthy that if an asymmetric object is seen as a symmetric object depending on the camera viewpoint, as shown in (d) and (e), our method estimates the 6D object pose exploiting valid primitive images. In addition, we observe that for the occlusion, if patches that determine the directionality of an object are visible, then our method reconstructs primitive images as shown in (c) and (j). As can be seen from these examples, PrimA6D++ can discern the case of object ambiguity through uncertainty estimation. Thanks to this superiority, the proposed method performs better than other RGB-based single view methods on the YCB-VIDEO dataset with several cases of occlusions and the T-LESS dataset with multiple symmetrical objects, as shown in \tabref{tab:ycb} and \tabref{tab:tless_bop}.

\subsubsection{Uncertainty for Object Pose Factor}

In our approach, the uncertainties of three primitive images are combined with the uncertainty for the final object pose factor. This section evaluates the relation between the object pose errors and the estimated uncertainties.

As shown in the last column of \figref{fig:scatter_plot}, we present the relation between the average distance errors of 6D object pose estimation and the norm $\left\| \boldsymbol{\sigma} \right\|$ of valid uncertainties. In the asymmetrical object, an increase in average distance errors yields an increase in the uncertainties. In the symmetrical object, we find a slight stagnation in some areas due to the ambiguity of the symmetrical object; afterwards, the trend line increases monotonically. This aspect is similar to the $\sigma_{z}$ graph of \figref{fig:scatter_plot}. We can conclude that the norm $\left\| \boldsymbol{\sigma} \right\|$ of valid uncertainties positively correlates with the 6D object pose errors.

\subsection{Results on the public datasets}

\subsubsection{YCB-Video dataset}

We evaluate the performance of our method with other \ac{SOTA} methods on the YCB-Video dataset. The qualitative results for the YCB-Video dataset are presented in \figref{fig:qualitative}. As shown in \tabref{tab:ycb}, PrimA6D++ shows the best performance in RGB-based single-view methods as well as the multi-view method compared to other methods. Since the proposed method deals with the ambiguity of object pose through uncertainty estimation, we obtain superior performance on the YCB-V dataset with various challenging points, including occlusion and clutter. Through graph optimization, our method outperforms all other methods in terms of the ADD(-S) metric.

\subsubsection{T-LESS dataset}

PrimA6D++ outperforms others with graph optimization in terms of $e_\mathrm{vsd}$ (\tabref{tab:tless_vsd}). Furthermore, we compare our method with those reported in the BOP challenge, as shown in \tabref{tab:tless_bop}. PrimA6D++ achieves the second highest performance in terms of $\mathrm{AR}$. However, it is worth noting that PrimA6D++ outperforms other methods in terms of $\mathrm{AR_{mspd}}$ which is relevant for AR applications and suitable for evaluating RGB-only methods. This is because PrimA6D++ estimates the 6D object pose by extracting the primitive axes coincident with the dominant axes of the objects visible in the images. In addition, with the graph optimization, PrimA6D++ shows a significant performance improvement over all of the metrics, especially $\mathrm{AR_{mssd}}$ which is deeply related to the chance of a successful robotic grasping. We conjecture that with the graph optimization of sequential data, the comprehension of the target workspace in 3D spaces gradually increases, which leads to improved accuracy. As a result, we achieve object pose accuracy sufficient for robot manipulation. The qualitative results for the T-LESS dataset are presented in \figref{fig:qualitative}.


\subsubsection{Effect of ICP refinement}

Before adding the estimated object pose proposals, we perform \ac{ICP} refinement using depth measurement for the cases where the norm $\left\| {\boldsymbol{\sigma}} \right\|$ of valid uncertainties is lower than a threshold $\sigma_{icp}$. In this section, we evaluate the effect of \ac{ICP} refinement on the performance of graph optimization. \tabref{tab:ycb} and \tabref{tab:tless_bop} list the performance variation depending on the presence of \ac{ICP} refinement of PrimA6D++ with graph optimization. Even without \ac{ICP} refinement, PrimA6D++ with graph optimization creates an exorbitant performance improvement since it guarantees global consistency of multi-view. \ac{ICP} refinement on the object pose proposals further improves accuracy.


\subsection{Real-Time Full System Demo}

For the real-time full system demonstration, we use the robot system as shown in \figref{fig:overview}. Since one network is trained to handle one object, the inference is performed in parallel using four GPUs to account for multiple objects. To recognize the objects in the workspace, the robot manipulator performs a scan around the objects; concurrently, the multi-object pose estimation is executed. In the end, with our proposed method, we demonstrate real-time scene recognition capability for visually-assisted robot manipulation. The experimental results are available in the supplementary video materials \texttt{prima6d.mp4}.


\section{Conclusion}
\label{sec:conclusion}

In this paper, we introduced a pipeline for multi-object pose estimation, which incorporates our novel ambiguity-aware 6D object pose estimation network. We devised a network to reconstruct the three rotation axis primitive images separately and infer the underlying uncertainty along each axis. Afterwards, leveraging the estimated uncertainty, we optimized these visual measurements and camera poses through a factor graph optimization. In the experiment, we validated the effect of ambiguity awareness using estimated uncertainties on the 6D object pose estimation, as well as the correlation between the estimated uncertainties and object pose errors. We demonstrated that the accuracy of multi-object pose estimation was improved through graph optimization. We further laid the groundwork for visually-assisted robot manipulation through real-time scene recognition.

\small
\bibliographystyle{IEEEtranN}
\bibliography{string-short,reference}

\end{document}